\documentclass[runningheads]{llncs} %

\usepackage[utf8]{inputenc}
\usepackage[T1]{fontenc}
\usepackage[english]{babel}
\usepackage{microtype}      %
\usepackage{csquotes}

\usepackage{amsmath}
\usepackage{amssymb}
\usepackage{bm}
\usepackage{booktabs}
\usepackage{xparse}

\usepackage[pdftex]{graphicx}
\usepackage{hyperref}

\renewcommand{\bbbr}{\ensuremath\mathbb{R}}

\DeclareDocumentCommand{\rscore}{O{f} O{i} m}{%
\ensuremath%
s_{#2}({#3}\,|\,{#1})}

\DeclareDocumentCommand{\rbscore}{O{f} O{i} m}{%
\ensuremath%
\bar{s}_{#2}({#3}\,|\,{#1})}

\DeclareDocumentCommand{\rtildescore}{O{f} O{i} m}{%
\ensuremath%
\tilde{s}_{#2}({#3}\,|\,{#1})}

\begin{document} %
\title{Scalable, Axiomatic Explanations of Deep Alzheimer's Diagnosis from Heterogeneous Data}
\titlerunning{Scalable, Axiomatic Explanations}
\author{Sebastian P{\"{o}}lsterl \and
   Christina Aigner \and
   Christian Wachinger} %
\authorrunning{S.~P{\"{o}}lsterl, C.~Aigner and C.~Wachinger} %

\institute{Artificial Intelligence in Medical Imaging (AI-Med),\\
   Department of Child and Adolescent Psychiatry,\\
   Ludwig-Maximilians-Universit{\"{a}}t, Munich, Germany%
   } %
\maketitle              %
\begin{abstract}
Deep Neural Networks (DNNs) have an enormous potential to learn
from complex biomedical data.
In particular, DNNs have been used to seamlessly
fuse heterogeneous information from neuroanatomy, genetics,
biomarkers, and neuropsychological tests
for highly accurate Alzheimer's disease diagnosis.
On the other hand, their black-box nature is still a barrier for the adoption of
such a system in the clinic, where interpretability is absolutely essential.
We propose Shapley Value Explanation of Heterogeneous Neural Networks
(SVEHNN) for explaining the Alzheimer's diagnosis
made by a DNN from the 3D point cloud of the neuroanatomy and tabular
biomarkers.
Our explanations are based on the Shapley value, which is the unique
method that satisfies all fundamental axioms
for local explanations previously established in the literature.
Thus, SVEHNN has many desirable characteristics
that previous work on interpretability for medical decision making is lacking.
To avoid the exponential time complexity of the Shapley value,
we propose to transform a given DNN into
a Lightweight Probabilistic Deep Network without re-training, thus achieving
a complexity only quadratic in the number of features.
In our experiments on synthetic and real data,
we show that we can closely approximate the exact Shapley value
with a dramatically reduced runtime
and can reveal the hidden knowledge the network has learned from the data.
\end{abstract} %
\section{Introduction}

In recent years, deep learning methods have become ubiquitous
across a wide range of applications in biomedicine
(see e.g. \cite{Ching2018} for an overview).
While these models may obtain a high predictive performance in
a given task, a major obstacle for their routine use in
the clinic is their black-box nature:
the inner workings that lead to a particular prediction remain
opaque to the user.
Methods that try to open the black-box and make its decision process
interpretable to the user, can be classified
as explainable AI~\cite{Arrieta2020}.
The goals of explainable AI are diverse and can be categorized
into five groups~\cite{Arrieta2020}:
to establish trust, discover causal relationships among inputs,
guarantee fairness and privacy, and
inform the user about the decision making process.

The focus of this paper is on \emph{local explanation}:
the ability to explain to the user how the prediction
for one specific input came to be.
We consider the input data to be heterogeneous,
namely a combination of
tabular biomarkers
and brain morphology,
both of which are common predictors for
various neurodegenerative diseases.
The shape of brain structures is rich in information
and preferred over simple summary statistics such as volume,
as demonstrated in recent deep learning approaches for
neuroanatomical shape analysis~\cite{GutierrezBecker2018,Kopper2020,Poelsterl2019}.
However, due to the non-Euclidean geometry,
deep learning on shapes requires dedicated network architectures
that differ substantially from standard convolutional
neural networks.
When also incorporating tabular biomarkers,
the network usually consists of two arms, each specific
to one particular input type.
Such networks can achieve impressive predictive performance,
but to consider their deployment in the clinic,
we need a principled methodology capable of
explaining their predictions.

We require
explanations to be easy to understand by a human
and to accurately imitate the deep learning model.
We attain these two properties by building
upon the strong theoretical guarantees of
the Shapley value~\cite{Shapley1953},
a quantity from cooperative game theory.
The work in \cite{Ancona2019,Lundberg2017,Sundararajan2019,Sundararajan2017}
established a set of fundamental axioms
local explanations ought to satisfy, and proved
that the Shapley value is the unique procedure that satisfies
(1) completeness, (2) null player, (3) symmetry, (4) linearity,
(5) continuity, (6) monotonicity, and (7) scale invariance.
Unfortunately, computing the Shapley value scales
exponential in the number of features, which
necessitates developing an efficient approximation.
To the best of our knowledge, the Shapley value
has not been used to explain predictions
made by a deep neural network from neuroanatomical shape
and clinical information before.

In this work, we propose Shapley Value Explanation of Heterogeneous Neural Networks
(SVEHNN): a scalable, axiomatic approach for explaining
the decision of a deep neural network integrating
neuroanatomical shape and tabular biomarkers.
SVEHNN is based on the Shapley value~\cite{Shapley1953}
and thus inherits its strong theoretic guarantees.
To overcome the cost, exponential in the number of features,
we efficiently approximate it by
converting standard layers
into corresponding probabilistic layers without re-training.
Thus, the computational complexity of explaining
a model's prediction is quadratic instead of exponential
in the number of features.
We assess the approximation error of SVEHNN on
synthetic data, for which the exact Shapley value can be computed,
and demonstrate its ability to explain Alzheimer's disease
diagnosis of patients on real clinical data.

\subsubsection{Related work.}
Early work on local explanations includes
the occlusion method, which selectively occludes portions of
the input and measures the change in the network's output~\cite{Zeiler2014}.
It has been employed in~\cite{GutierrezBecker2018}
for Alzheimer's disease (AD) diagnosis based on
shape representations of multiple brain structures, and
in~\cite{Li2018} for
autism spectrum disorder (ASD) diagnosis based on fMRI.
A second group of algorithms obtains local explanations
by back-propagating a network's output
through each layer of the network until it reaches
individual features of the input~\cite{Bach2015,Selvaraju2017,Shrikumar2017,Sundararajan2017,Zhao2018}.
They differ in how the gradients are incorporated
into the final feature relevance score.
In \cite{Zhao2018}, the authors incorporate
forward activations and gradients in the relevance
score and applied it to classification of cellular electron cryo-tomography.
Different from the above,
Zhuang~et~al.~\cite{Zhuang2019} proposed an invertible network
for ASD diagnosis.
By projecting the latent representation
of an input on the decision boundary of the last linear layer
and inverting this point, they obtain the corresponding
representation in the input space.
As shown in~\cite{Ancona2019,Lundberg2017,Sundararajan2019,Sundararajan2017},
the vast majority of the methods above do not satisfy
all fundamental axioms of local explanations outlined above.
Most relevant to our work is
the efficient approximation of the
Shapley value for multi-layer perceptrons and
convolutional neural networks in~\cite{Ancona2019}. However, it is not
applicable to heterogeneous inputs we consider here.
In fact, most methods were developed for image-based
convolutional neural networks, and it remains unclear how they
can be used for non-image data, e.g.,
when defining a suitable ``non-informative reference input''
for gradient-based methods~\cite{Bach2015,Selvaraju2017,Shrikumar2017,Sundararajan2017,Zhao2018}.

\section{Methods}

We consider explaining predictions
made by a Wide and Deep PointNet (WDPN, \cite{Poelsterl2019}) from the hippocampus
shape and clinical variables of an individual for AD
diagnosis by estimating the influence each input feature has on
the model's output.
Before describing our contribution,
we will first briefly summarize the network's architecture and
define the Shapley value.

\subsection{Wide and Deep PointNet}

\begin{figure}[tb]
	\centering
	\includegraphics[scale=1]{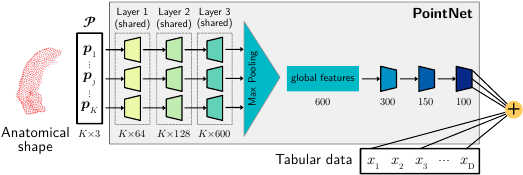}
	\caption{\label{fig:pointnet}%
	Wide and Deep PointNet Architecture (adapted from \cite{Poelsterl2019}).}
\end{figure}

Figure~\ref{fig:pointnet} depicts the architecture of our network
to predict the probability of AD diagnosis.
We represent the hippocampus shape as a point cloud $\mathcal{P}$,
and known clinical variables
associated with AD as tabular data $\bm{x} \in \bbbr^D$~\cite{Poelsterl2019}.
Point clouds are fed to a PointNet~\cite{Qi2017}, which
first passes each coordinate vector
through a multilayer perceptron with shared weights
among all points,
before aggregating point descriptors
using max pooling.
We fuse $\bm{x}$ with the latent representation
of $\mathcal{P}$ in a final linear layer.
To account for non-linear effects, WDPN augments clinical markers
by B-spline expansion or interactions.

\subsection{Shapley Value}

Given a trained WDPN,
$f : \bbbr^{K \times 3} \times \bbbr^D \rightarrow [0; 1]$,
we want to explain the predicted probability of AD
for one particular input $\mathbf{z} = (\mathcal{P}, \bm{x})$
by estimating the contribution individual points
in the point cloud and clinical features have on the prediction.
In particular, we propose to estimate
the Shapley value~\cite{Shapley1953}, which we will define next.

We denote by $\mathcal{F}$ the set of all features comprising the input,
and $\mathbf{z}_\mathcal{S}$ the subset indexed by $\mathcal{S} \subseteq \mathcal{F}$.
Let $g: \mathbf{P}(\mathcal{F}) \rightarrow \bbbr$
be a set function defined on the power set of $\mathcal{F}$,
with $g(\mathcal{S}) \neq 0$, $g(\emptyset) = 0$.
The contribution of a set of features $\mathcal{S}$ on the prediction is given
by the set function
$
    g(\mathcal{S}) = f(\mathbf{z}_\mathcal{S};
    \mathbf{z}_{\mathcal{F}\setminus\mathcal{S}}^\text{bl})
    - f(\mathbf{z}^\text{bl}) ,
$
where the first term corresponds to the prediction of the model
after replacing all features
not in $\mathcal{S}$ with a baseline value from the vector
$\mathbf{z}^\text{bl}$.
The Shapley value for feature $i$ is defined as
\begin{align}\label{eq:shapley-value}
    \rscore{\mathbf{z}} = {\textstyle\frac{1}{|\mathcal{F}|!}
    \sum_{\mathcal{S} \subseteq \mathcal{F}\backslash\{i\}}
    |\mathcal{S}|! \cdot (|\mathcal{F}| - |\mathcal{S}| - 1)!
    \left( g(\mathcal{S} \cup \{i\}) - g(\mathcal{S}) \right) .}
\end{align}

\paragraph{Approximate Shapley Value.}

Computing~\eqref{eq:shapley-value} exactly would require exponential
many evaluations of the difference
$\Delta_i = g(\mathcal{S} \cup \{i\}) - g(\mathcal{S})$
for each feature $i$.
Therefore, we employ the approximate Shapley value, first proposed for
voting games in~\cite{Fatima2008}.
Let $\mathbb{E}_k(\Delta_i)$ denote the marginal contribution
of feature $i$, where
the expectation is over all $\mathcal{S}$ with $|\mathcal{S}| = k$.
By writing \eqref{eq:shapley-value} in an alternative form
where we explicitly sum over all sets $\mathcal{S}$ of equal size
and noting that there are ${|\mathcal{F}| - 1} \choose {k}$ possible
sets of size $k$, we arrive at the
approximate Shapley value $\bar{s}_i$:
\begin{align}
    \label{eq:approx-shapley}
    \rscore{\mathbf{z}} &= \frac{1}{|\mathcal{F}|!}
    \sum_{k=0}^{|\mathcal{F}| - 1}
    \sum_{\substack{
    \mathcal{S} \subseteq \mathcal{F}\backslash\{i\}\\
    |\mathcal{S}| = k}}
    k! (|\mathcal{F}| - k - 1)!
    \cdot \Delta_i
    \approx
    \frac{1}{|\mathcal{F}|} \sum_{k=0}^{|\mathcal{F}|-1} \mathbb{E}_k(\Delta_i)
    = \rbscore{\mathbf{z}} .
\end{align}
We will now focus on our main contribution to efficiently estimate the expectation
\begin{equation}\label{eq:shapley-difference}
    \mathbb{E}_k(\Delta_i) = \mathbb{E}_k[
    f(\mathbf{z}_{\mathcal{S} \cup \{i\}};
      \mathbf{z}_{\mathcal{F}\setminus\mathcal{S} \cup \{i\}}^\text{bl})
    ]
    - \mathbb{E}_k[
        f(\mathbf{z}_\mathcal{S};
        \mathbf{z}_{\mathcal{F}\setminus\mathcal{S}}^\text{bl})
    ] .
\end{equation}

\subsection{Efficient Estimation of the Approximate Shapley Value}

Our main contribution is based on the observation that
we can treat $\mathbf{z}_\mathcal{S}$ over all sets $\mathcal{S}$
of size $k$ as a source of aleatoric uncertainty, i.e.,
we are unsure about the input~\cite{Ancona2019}.
Then, the objective becomes propagating the aleatoric uncertainty
through the network.
This is a challenging task, because we need to create a version
of the WDPN that is probabilistic and faithful to the original model.
To this end, we propose a novel probabilistic WDPN, inspired by
the Lightweight Probabilistic Deep Network~(LPDN, \cite{Gast2018}),
that models aleatoric uncertainty by assuming inputs
are comprised of independent univariate normal distributions,
one for each feature.
The resulting probabilistic WDPN
directly outputs an estimate of
$\mathbb{E}_k [
    f(\mathbf{z}_\mathcal{S};
    \mathbf{z}_{\mathcal{F}\setminus\mathcal{S}}^\text{bl})
]$
for a fixed $k$.
Next, we will propose a new set of layers to transform
inputs into distributions over sets $\mathcal{S}$
of size $k$.

\paragraph{Probabilistic PointNet.}

The first part of the network in Fig.~\ref{fig:pointnet}
processes the point cloud. It consists of multiple fully-connected
layers with ReLU activation and batch normalization, where
weights are shared across points.
For the initial linear layer with weights $\mathbf{W}$,
the $m$-th output for the $j$-th point
is $h_{jm} = \sum_{l=1}^3 p_{jl} W_{lm}$.
To make it probabilistic, we need to account for
$\mathbf{p}_j$ being included in $\mathcal{S}$ randomly.
Since each $\mathcal{S}$ is selected with probability
${{|\mathcal{F}|} \choose {k}}^{-1}$ and there are
${{|\mathcal{F}|-1} \choose {k-1}}$ sets $S$ containing $\mathbf{p}_j$,
we have
\begin{equation}\label{eq:sampling-mean}
    \mathbb{E}_k[h_{jm}]
    = \sum_{\substack{
        \mathcal{S} \subseteq \mathcal{F}\\
        |\mathcal{S}| = k}}
        {{|\mathcal{F}|} \choose {k}}^{-1}
        {{|\mathcal{F}|-1} \choose {k-1}}
        h_{jm}
    = \frac{k}{|\mathcal{F}|}
    h_{jm} .
\end{equation}
This is a well-known result from sampling theory~\cite{Ancona2019,Cochran1977}, which also tells us
that $h_{jm}$ can be approximated with a normal distribution
with mean \eqref{eq:sampling-mean} and variance
\begin{equation}\label{eq:sampling-variance}
    \mathbb{V}_k(h_{jm}) =
    k \frac{|\mathcal{F}| - k}{|\mathcal{F}| - 1} \left[
        \frac{1}{|\mathcal{F}|} \sum_{l=1}^3 (p_{jl} W_{lm})^2
        - \left( \frac{1}{|\mathcal{F}|} h_{jm} \right)^2
    \right] .
\end{equation}
After the first linear layer
each output unit is approximated by a normal distribution, and
we can employ a LPDN for all subsequent layers,
by replacing ReLU, batch-norm, and max-pooling with their
respective probabilistic versions~\cite{Gast2018}.
The final output of the probabilistic PointNet is a
global descriptor of the whole point cloud, where
each feature has a mean and a variance attached to it.

\paragraph{Probabilistic Wide and Deep PointNet.}

The final step in estimating the expectation \eqref{eq:shapley-difference}
is the integration of clinical markers with the latent point cloud
representation (yellow node in Fig.~\ref{fig:pointnet}).
Both information is combined in a linear layer, for which
we need to propagate uncertainty due to the distributions describing
the point cloud, and due to $\mathcal{S}$
covering a subset of clinical markers.
Since a linear layer is an associative operation, we can compute
the sum over the latent point cloud features and the clinical features
separately, before combining the respective results,
thus we split the expectation \eqref{eq:shapley-difference}
into $\mathbb{E}_k(\Delta_{i})
= \mathbb{E}_k(\Delta_{i}^{\bm{x}}) + \mathbb{E}_k(\Delta_{i}^{\mathcal{P}})$.
To compute the expectation with respect to tabular data $\bm{x}$,
we utilize that for a
linear layer with bias $b$ and weights $\mathbf{w} \in \bbbr^D$,
$f(\bm{x}) = b + \mathbf{w}^\top\bm{x}$,
thus $\mathbb{E}_k(\Delta_{i}^{\bm{x}}) = w_i x_i$, if feature $i$ is tabular and
$\mathbb{E}_k(\Delta_{i}^{\bm{x}}) = 0$ otherwise.
The uncertainty due to the latent point cloud features can
be propagated by a linear probabilistic layer~\cite{Gast2018}
with mean $\mu_{k}^{\mathcal{S}}$, yielding
$\mathbb{E}_k[
    f(\mathbf{z}_\mathcal{S};
    \mathbf{z}_{\mathcal{F}\setminus\mathcal{S}}^\text{bl})
] = \mu_{k}^{\mathcal{S}}$.
Doing the same, but accounting for the inclusion of $i$ in $\mathcal{S}$, we
have
$\mathbb{E}_k[
    f(\mathbf{z}_{\mathcal{S} \cup \{i\}};
      \mathbf{z}_{\mathcal{F}\setminus\mathcal{S} \cup \{i\}}^\text{bl})
    ] = \mu_{k}^{\mathcal{S} \cup \{i\}}
$.
Finally, we subtract both means yielding
$\mathbb{E}_k(\Delta_{i}^{\mathcal{P}}) =
\mu_{k}^{\mathcal{S} \cup \{i\}} - \mu_{k}^{\mathcal{S}}$.
As a result, we can estimate $\mathbb{E}_k(\Delta_{i})$
with only two forward-passes, while keeping
$\mathbf{z}$ and $k=|\mathcal{S}|$ fixed.

\paragraph{Estimating the Approximate Shapley Value.}
Equipped with an efficient way to estimate the expectation
over a fixed number of features in $\mathcal{S}$, we
can compute the difference in~\eqref{eq:shapley-difference}.
For the first term, we only need a single forward pass
through the probabilistic WDPN using the
original input with all features.
For the second term, we again need one forward pass, but
without the contribution of the $i$-th feature, which
is replaced by a baseline value. This value differs depending
on whether the $i$-th feature is a point in the hippocampus
or a clinical marker.
Thus, the complexity to estimate the approximate Shapley value
for all features in the input is $\mathcal{O}((D + |\mathcal{P}|)^2 )$.
For high-dimensional inputs, we can further reduce the complexity
by estimating the average in \eqref{eq:approx-shapley} via
$M$ Monte Carlo samples, yielding $\mathcal{O}(M (D + |\mathcal{P}|))$.

\paragraph{Choosing a Baseline.}
Selection of an appropriate baseline $\mathbf{z}^\text{bl}$ is crucial
in obtaining semantically meaningful explanations, as the $i$-th
Shapley value reflects the contribution of feature $i$ to
the difference $f(\mathbf{z}) - f(\mathbf{z}^\text{bl})
= \sum_{i=1}^{|\mathcal{F}|} \rscore{\mathbf{z}}$~\cite{Shapley1953}.
Assuming no clinical marker has been zero-encoded, we
can replace the original value with zero to eliminate its
impact.
If the $i$-th feature belongs to the point cloud, the situation
is more nuanced.
Using all-zeros would describe
a point at the origin and Shapley values would explain
the difference to the prediction made from the origin,
which has no semantic meaning in the case of the hippocampus.
We propose a better alternative by
replacing it with a matching point from a hull containing
all point clouds in the dataset.
Our motivation for this approach is that it
avoids erratic changes of the point cloud's surface when
inserting a point at the origin, and that it
provides a common, semantically meaningful reference
for all patients, which eases interpretation of
explanations across multiple patients.

\section{Experiments}
In this section, we are evaluating the computational efficiency,
approximation error, and semantics of SVEHNN compared to the exact Shapley value~\cite{Shapley1953},
Occlusion~\cite{Zeiler2014}, and Shapley sampling~\cite{Castro2009}.
Note that except for Occlusion, these methods have not been
used to explain predictions from shape data before.
For each method, we compare three different approaches to choosing the baseline
value in point clouds: (i) moving a point to the origin (zero),
and (ii) replacing a point by its matching point from the common hull (hull).
We quantitatively assess differences with respect to the exact Shapley value
in terms of mean squared error (MSE), Spearman rank correlation (SRC),
and normalized discounted cumulative gain (NDCG,~\cite{Jaervelin2002}).
MSE captures the overall approximation error,
SRC is a measure of agreement between rankings of features,
and NDCG is similar to SRC, but penalizes errors in important features
-- according to the ground truth -- more.
For NDCG, we use the absolute Shapley value as importance measure.
We run Shapley sampling~\cite{Castro2009} with $M=2,000$ Monte Carlo samples,
which results in a runtime of
$\mathcal{O}(|\mathcal{P}| M)$.
Note that sampling usually requires $M \gg |\mathcal{P}|$ for a good approximation.

\subsection{Synthetic Shape Data Set} %
To assess the approximation error compared to the exact Shapley value
in a reasonable amount of time, we created a synthetic dataset
of point clouds comprising 16 points ($|\mathcal{P}| = 16$).
This allows calculation of the exact Shapley values across all
$2^{16-1}$ subsets of $\mathcal{P}$ and use it as ground truth.
We use a binary PointNet classifier
to distinguish point clouds of the characters `X' and `I'.

\begin{table}[tb]
\centering
\caption{\label{tab:results-synthetic}%
    Mean difference to exact Shapley value across 100 synthetic point clouds with zero baseline.
    NE:~Number of network evaluations for each example.}
\begin{scriptsize}
\begin{tabular}{l @{~~} r @{~~} r @{~~} r @{~~} r}
    \toprule
\multicolumn{1}{c}{Method}
& \multicolumn{1}{c}{MSE}
& \multicolumn{1}{c}{SRC}
& \multicolumn{1}{c}{NDCG}
& \multicolumn{1}{c}{NE} \\
\midrule
Exact     & 0        & 1       & 1       & 65,538 \\
Sampling  &   0.0008 &  0.9505 &  0.9986 & 32,000 \\
Sampling  &   0.0340 &  0.5440 &  0.9481 & 512 \\
Occlusion &  16.1311 &  0.3180 &  0.8659 & 17 \\
SVEHNN    &   0.0443 &  0.6918 &  0.9641 & 512 \\
\bottomrule
\end{tabular}
\end{scriptsize}
\end{table}

Results in Table~\ref{tab:results-synthetic} show
that occlusion fails in all categories, except runtime efficiency.
SVEHNN closely approximates the exact Shapley value
in terms of MSE with less than 0.8\% of network evaluations.
Running Shapley sampling until convergence (second row) ranks slightly
higher, but at more than 60 times the cost of SVEHNN, it is a high
cost for an improvement of less than 0.035 in NDCG.
When considering Shapley sampling with the same runtime
as SVEHNN (third row), rank metrics drop below that of SVEHNN;
the MSE remains lower, which is due to the Gaussian approximation
of activations in probabilistic layers of SVEHNN.
Overall, we can conclude that SVEHNN is computationally efficient
with only a small loss in accuracy.

\subsection{Alzheimer's Disease Diagnosis}
In this experiment, we use data from the
Alzheimer's Disease Neuroimaging Initiative~\cite{Jack2008}.
Brain scans were processed
with FreeSurfer~\cite{Fischl2012} from which we obtained the surface
of the left hippocampus,
represented as a point cloud with 1024 points.
For tabular clinical data, we use age, gender, education (in orthogonal polynomial coding), APOE4,
FDG-PET, AV45-PET, A$\beta_{42}$,
total tau (t-tau), and
phosphorylated tau (p-tau).
We account for non-linear age effects using
a natural B-spline expansion with four degrees of freedom.
We exclude patients diagnosed with mild cognitive impairment
and train a WDPN~\cite{Poelsterl2019}
to tell healthy controls from patients with AD.
Data is split into 1308 visits for training, 169
for hyper-parameter tuning, and 176 for testing;
all visits from one subject are included in the same split.
The WDPN achieves a balanced accuracy of 0.942 on the test data.
We run SVEHNN with 150 Monte Carlo samples and the hull baseline.

\begin{figure}[tb]
\begin{minipage}[t]{.49\textwidth}
    \centering
    \includegraphics[scale=.27]{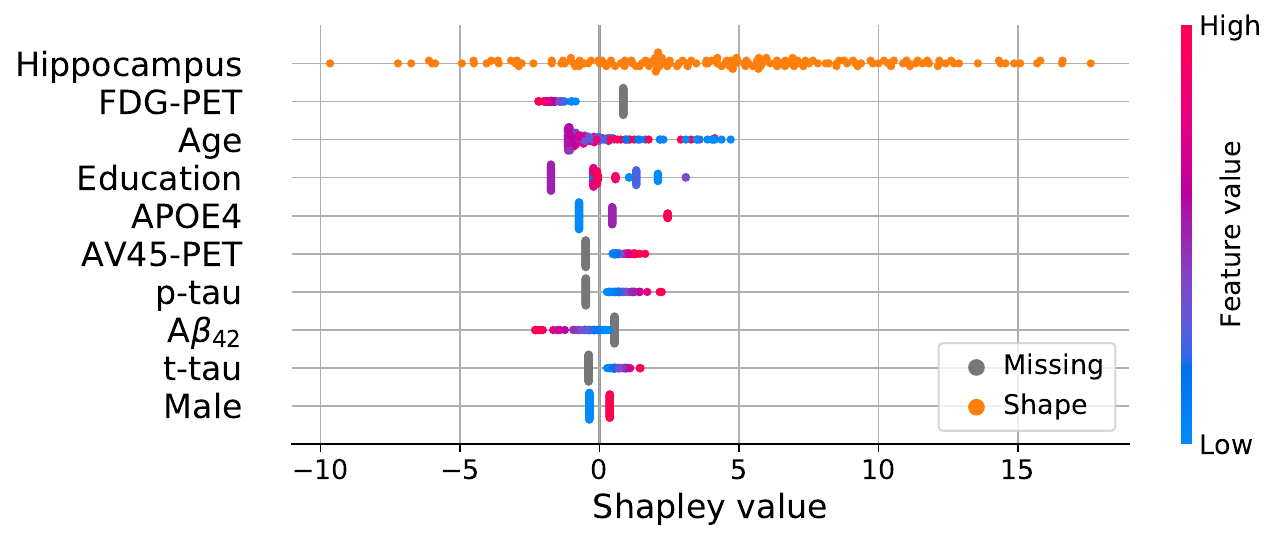}
    \caption{\label{fig:adni-shapley-testset}%
    Shapley values of biomarkers and hippocampus shape of 167 correctly classified patients
    (dots, 110 healthy, 57 AD), sorted by mean relevance.}
\end{minipage}\quad%
\begin{minipage}[t]{.49\textwidth}
    \centering%
    \includegraphics[scale=.29]{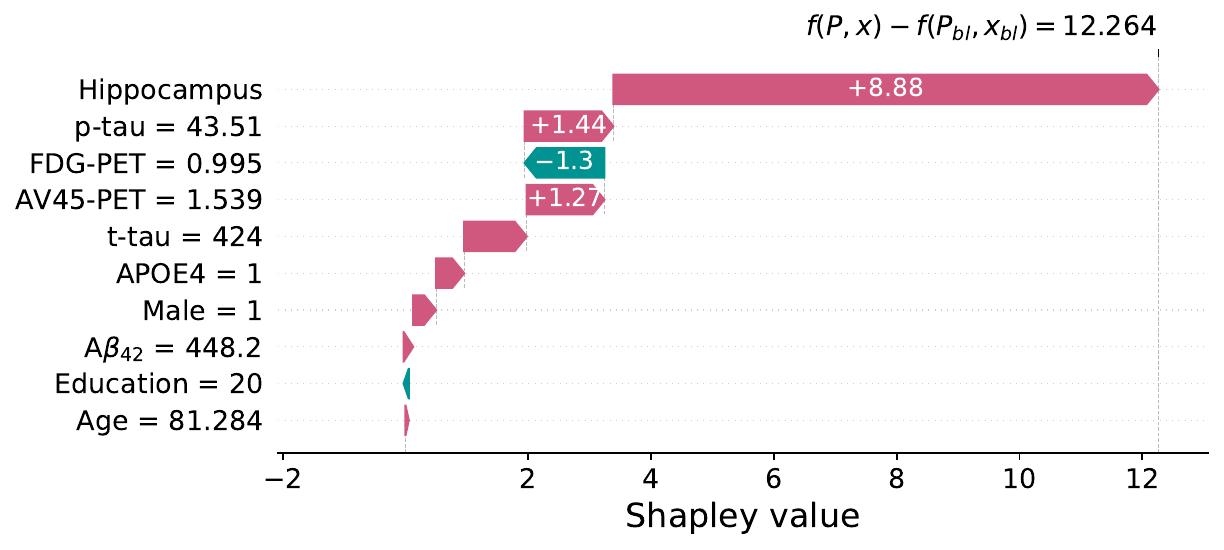}%
    \caption{\label{fig:waterfall-ad}%
    Shapley values of biomarkers of a single AD patient.}
\end{minipage}
\end{figure}

\begin{figure}[tb]
    \centering
    \includegraphics[height=0.18\textheight]{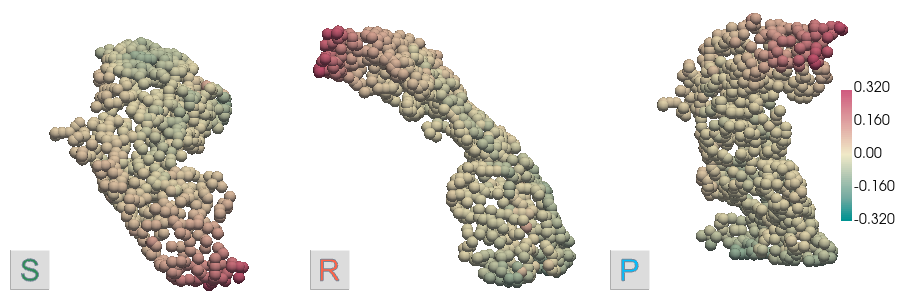}
    \caption{\label{fig:pointcloud}%
    Shapley values of hippocampus shape of same patient as
    in Fig.~\ref{fig:waterfall-ad}.
    Left: Superior view.
    Middle: Right view. Right: Posterior view.}
\end{figure}

Figure~\ref{fig:adni-shapley-testset} depicts the Shapley values
of individual patients and features for all correctly classified test patients,
sorted by the average feature relevance.
For tabular biomarkers, colors indicate the range of feature values
from low (blue) to high (red). For instance, APOE4 Shapley values of
patients with no APOE4 allele are depicted in blue, with one allele
in purple, and with two alleles in red.
It shows that for a few patients the hippocampus shape is
very important, whereas for most patients its relevance is similar
to that of tabular biomarkers.
Overall, relevance scores echo clinically validated results:
high concentrations of p-tau/t-tau and APOE4
are markers for AD, whereas
low levels of $A\beta_{42}$ and
many years of education are protective~\cite{Blennow2001,Genin2011,Meng2012}.
To provide an example of a local explanation by SVEHNN,
we selected one AD patient (see Fig.~\ref{fig:waterfall-ad}).
It shows that the AD diagnosis was driven by the hippocampus shape,
followed by p-tau, FDG-PET, and AV45-PET.
The explanation of the hippocampus reveals that
points located in the CA1 subfield were most important for diagnosis
(see Fig.~\ref{fig:pointcloud}).
This is reassuring, because atrophy in the CA1 subfield
is an established marker for AD~\cite{Renaud2013}.
By cross-referencing the explanation with clinical knowledge,
we can conclude that the predicted AD diagnosis is likely
trustworthy.

\section{Conclusion}

Obtaining comprehensible and faithful explanations of
decisions made by deep neural networks are paramount for
the deployment of such systems in the clinic.
We proposed a principled methodology for explaining individual
predictions, which can help to build trust among the users of such a system.
In this work, we studied explaining networks integrating
neuroanatomical shape and tabular biomarkers, which
has not been studied before.
We followed an axiomatic approach based on the Shapley value -- the unique procedure
that satisfies all fundamental axioms of local explanation methods.
Our proposed method Shapley Value Explanation of Heterogeneous Neural Networks
(SVEHNN) closely approximates the exact Shapley value
while requiring only a quadratic instead of exponential number
of network evaluations.
Finally, we illustrated how SVEHNN can help to understand
Alzheimer's diagnosis made by a network from heterogeneous data
to reveal the hidden knowledge the network has learned from the data.

\subsubsection*{Acknowledgements.}
This research was supported by the Bavarian State Ministry of Science and the Arts and coordinated by the Bavarian Research Institute for Digital Transformation,
and the Federal Ministry of Education and Research in the call for Computational Life Sciences (DeepMentia, 031L0200A).
\bibliographystyle{splncs04}
\bibliography{references}

\end{document}